\documentclass{article}

     \PassOptionsToPackage{numbers, compress}{natbib}

    \usepackage[preprint]{neurips_2024}

\usepackage[utf8]{inputenc} 
\usepackage[T1]{fontenc}    
\usepackage{hyperref}       
\usepackage{url}            
\usepackage{booktabs}       
\usepackage{amsfonts}       
\usepackage{nicefrac}       
\usepackage{microtype}      
\usepackage{xcolor}         
\usepackage{multicol}
\usepackage{multirow}
\usepackage{enumitem}
\usepackage{bm}
\usepackage{graphicx}
\usepackage{subcaption}
\usepackage{amsmath}
\usepackage{listings}
\usepackage{pifont}

\setlist[itemize]{leftmargin=*}
\setlist[enumerate]{leftmargin=*}

\definecolor{codegreen}{rgb}{0,0.6,0}
\definecolor{codegray}{rgb}{0.5,0.5,0.5}
\definecolor{codepurple}{rgb}{0.58,0,0.82}
\definecolor{backcolour}{rgb}{0.95,0.95,0.95}
\lstdefinestyle{mystyle}{
  backgroundcolor=\color{backcolour}, commentstyle=\color{codegreen},
  keywordstyle=\color{magenta},
  numberstyle=\tiny\color{codegray},
  stringstyle=\color{codepurple},
  basicstyle=\ttfamily\footnotesize,
  breakatwhitespace=true,         
  breaklines=true,       
  breakindent=0pt,
  captionpos=b,                    
  keepspaces=true,                 
  numbers=none,                    
  numbersep=5pt,                  
  showspaces=false,                
  showstringspaces=false,
  showtabs=false,                  
  tabsize=2,
  extendedchars=true, 
  literate=%
  {Ö}{{\"O}}1
  {Ä}{{\"A}}1
  {Å}{{\AA{}}}1
  {Ü}{{\"U}}1
  {ß}{{\ss}}1
  {ü}{{\"u}}1
  {ö}{{\"o}}1
  {ä}{{\"a}}1
  {å}{{\aa{}}}1
  {á}{{\'a}}1
  {ã}{{\~a}}1
  {é}{{\'e}}1,
}
\title{AutoFlow: Automated Workflow Generation for \\Large Language Model Agents}

\author{%
  Zelong Li \\ 
  Rutgers University \\
  \texttt{zelong.li@rutgers.edu} \\
  \And 
  Shuyuan Xu \\
  Rutgers University \\
  \texttt{shuyuan.xu@rutgers.edu} \\
  \And 
  Kai Mei \\
  Rutgers University \\
  \texttt{kai.mei@rutgers.edu} \\
  \And
  Wenyue Hua \\
  Rutgers University \\
  \texttt{wenyue.hua@rutgers.edu}\\
  \And
  Balaji Rama \\
  Independent Researcher \\
  \texttt{balajirw10@gmail.com}\\
  \And
  Om Raheja \\
  Independent Researcher \\
  \texttt{omr@rslp.org}
  \And
  Hao Wang \\
  Rutgers University \\
  \texttt{hw488@cs.rutgers.edu} \\ 
  \And
  He Zhu \\
  Rutgers University \\
  \texttt{hz375@cs.rutgers.edu} \\ 
  \And
  Yongfeng Zhang \\
  Rutgers University \\
  \texttt{yongfeng.zhang@rutgers.edu} \\ 
}

\begin{document}

\maketitle

\begin{abstract}
Recent advancements in Large Language Models (LLMs) have shown significant progress in understanding complex natural language. 
One important application of LLM is LLM-based AI Agent, which leverages the ability of LLM as well as external tools for complex-task solving.
To make sure LLM Agents follow an effective and reliable procedure to solve the given task, manually designed workflows are usually used to guide the working mechanism of agents. 
However, manually designing the workflows requires considerable efforts and domain knowledge, making it difficult to develop and deploy agents on massive scales.
To address these issues, we propose AutoFlow, a framework designed to automatically generate workflows for agents to 
solve complex tasks.
AutoFlow takes natural language program as the format of agent workflow and employs a workflow optimization procedure to iteratively optimize the workflow quality. Besides, this work offers two workflow generation methods: fine-tuning-based and in-context-based methods, making the AutoFlow framework applicable to both open-source and closed-source LLMs. 
Experimental results show that our framework can produce robust and reliable agent workflows.
We believe that the automatic generation and interpretation of workflows in natural language represent a promising paradigm for solving complex tasks, particularly with the rapid development of LLMs.
The source code of this work is available at \url{https://github.com/agiresearch/AutoFlow}.
\end{abstract}

\section{Introduction}

Recent advancements in Large Language Models (LLMs) have demonstrated substantial progress in understanding and processing complex natural language. These developments have opened up a wide array of applications, among which the deployment of LLM-based AI agents stands out. These agents leverage the capabilities of LLMs along with external tools to tackle intricate tasks, ranging from data analysis \citep{openagi}, software development \citep{li2023camel, qian2023communicative}, scientific research \citep{boiko2023emergent}, travel planning \citep{xie2024travelplanner} to many other decision-making processes in various domains.

One of the critical aspects of ensuring that LLM-based AI agents operate effectively and reliably is the design of workflows that guide their task-solving procedures. For example, an LLM-based agent for fake news detection may execute under the following workflow designed by information and communication experts \cite{li2024large}: 1) Check the URL, 2) Check the language, 3) Commonsense evaluation, 4) Standpoint evaluation, 5) Summarize the findings, and 6) Classification. The agent executes the workflow step by step, and each step may call the LLM or external tools to gather useful information for the final summarization and classification.

Traditionally, these workflows are manually crafted, requiring significant effort and deep domain knowledge. This manual process poses a substantial barrier to the large-scale development and deployment of AI agents, as it is both time-consuming and resource-intensive. 

To address the challenges associated with manual workflow design, this paper proposes AutoFlow, a novel framework aimed at the automatic generation of workflows for AI agents to solve complex tasks. AutoFlow represents workflows in the form of natural language programs \cite{xu2024core}, facilitating easier comprehension and interaction. Central to AutoFlow is a workflow optimization procedure that iteratively refines the quality of the generated workflows, ensuring robustness and reliability.

\begin{figure}[t]
    \centering
    \includegraphics[width=0.9 \textwidth]{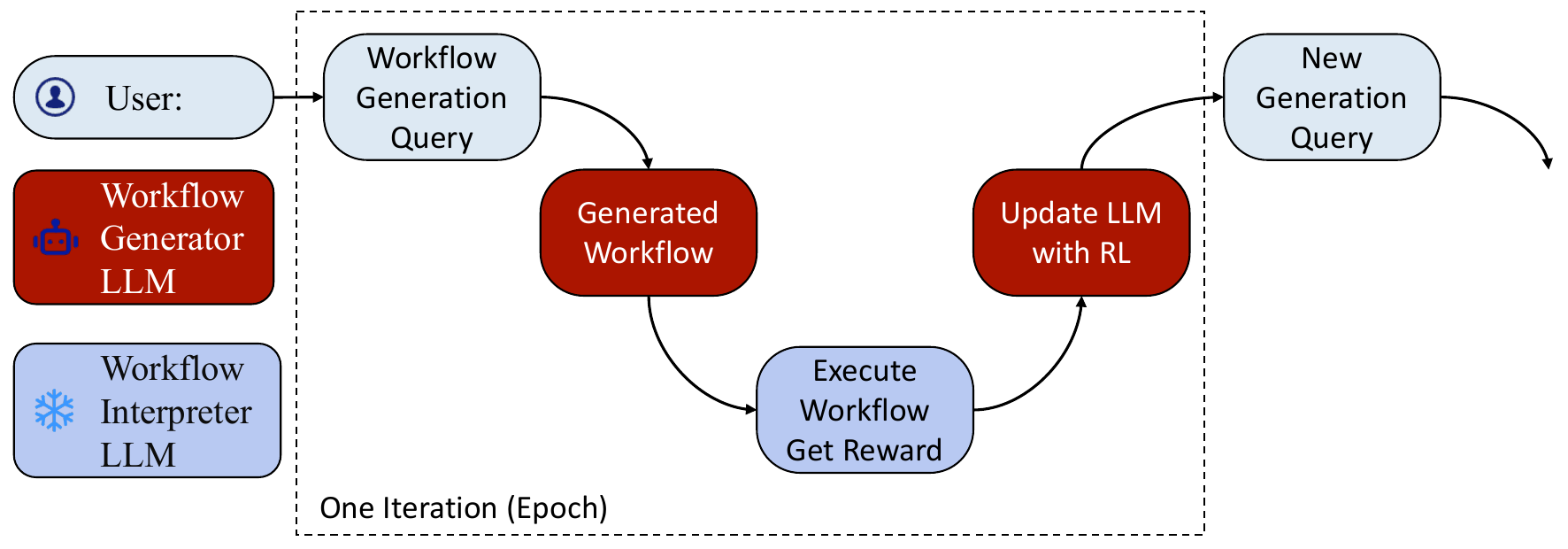}
    \caption{The overall generation process of AutoFlow using reinforcement learning reward for LLMs}
    \vspace{-15pt}
    \label{fig:workflow}
\end{figure}

Technically, AutoFlow introduces two innovative workflow generation methods: a fine-tuning-based method and an in-context-based method. The fine-tuning-based approach customizes the workflow generation process for specific tasks and domains by adjusting the parameters of the LLMs. In contrast, the in-context-based method utilizes contextual information to guide the generation process without the need for extensive fine-tuning, making it suitable for both open-source and closed-source LLMs.
More specifically, as shown in Figure \ref{fig:workflow}, the user will provide a workflow generation query to describe the type of tasks. Based on the query, the generator LLM generates a workflow and the frozen interpreter LLM executes the generated workflow on the dataset, with evaluating performance as the reward. Then, AutoFlow uses reinforcement learning (RL) to update the generator LLM with the reward. This process can be seen as one training iteration and the generator LLM expects to learn how to generate effective and optimal workflows after several iterations. 

Our experimental results validate the effectiveness of the AutoFlow framework, showing that the generated workflows by AutoFlow outperform manually designed ones while keeping readability, and showcasing its ability to produce high-quality workflows that enable AI agents to perform complex tasks with a high degree of reliability. The automatic generation and interpretation of workflows in natural language not only streamline the development process but also represent a promising paradigm for addressing complex problems, especially in the context of the rapid evolution of LLM technologies.
In summary, this paper makes the following contributions:
\begin{itemize}[leftmargin=*]
    \item We introduce AutoFlow, a framework that can automatically generate workflows in natural language so that the workflows can be precisely interpreted by LLMs while reducing human efforts.
    \item We propose two methods, the fine-tuning method and the in-context learning method, to incorporate RL in the workflow generation process for both open-source and closed-source LLMs.
    \item We conduct experiments through benchmark tasks to validate the AutoFlow framework, contributing to higher valid plan rates and overall performance while keeping the generated natural language workflow readable by humans.
\end{itemize}

In the following part of this paper, we first review the related work in Section \ref{sec:related_work}. In Section \ref{sec:background}, we introduce how to represent workflows in natural language and our motivations. In Section \ref{sec:autoflow}, we demonstrate the detailed design of our AutoFlow framework, including two learning methods, fine-tuning and in-context learning methods. We provide and analyze the experimental results on benchmark datasets in Section \ref{sec:experiment}, and finally conclude our work and suggest potential avenues for future research in Section \ref{sec:conclusions}.

\section{Related Work}
\label{sec:related_work}


\subsection{LLM Agents and Workflow}

AI agent is an autonomous entity capable of making decisions and executing actions in a given environment to effectively handle various complex tasks \citep{mei2024aios, ge2023llm, wang2023survey, xi2023rise}. Recently, with the rapid advancement of Large Language Models (LLMs), LLM-based AI agents have become an important type of agent for complex task solving \citep{openagi,li2023camel,qian2023communicative}, such as reasoning, planning and coding.

\textit{Reasoning}: LLMs typically break down complex tasks into a series of steps, constituting a chain of reasoning \cite{wei2022chain}. Approaches such as Chain of Thought (CoT) and its derivatives \cite{wei2022chain, kojima2022large}, including tree \cite{yao2024tree} and graph structures \cite{besta2024graph}, are commonly used. The self-consistency method \cite{wang2022self} samples multiple reasoning paths and selects the most consistent outcome through voting. 

\textit{Planning}: Planning tasks require LLMs to generate a sequence of actions to achieve specific goals \cite{hao2023reasoning}. Recent studies have designed platforms to test LLMs' planning abilities in areas such as expert model integration \citep{openagi}, travel task planning \citep{xie2024travelplanner}, and tool usage \citep{yuan2024easytool}. However, a known issue is that LLMs may generate non-executable, invalid or grammatically wrong plans, such as using a piece of text as input to an image-processing tool. To solve the problem, some studies \citep{openagi, yuan2024easytool} use post-processing method to extract a chain of tools from the generated texts, which use LLM itself as a parser to post-process the generated text. Further, recent attempts integrate finite state machines into LLMs to enhance human's controllability of LLM in planning \cite{li2024formalllm, wu2024stateflow}. The ReAct approach \cite{yao2022react} also uses external tools such as search engines to improve LLM planning. In this work, we build on these ideas to enhance the executability of the generated frameworks. 

\textit{Coding}: LLMs can generate code to solve complex tasks, reducing the need for manual programming \cite{lyu2023faithful, xu2024core, jojic2023gpt, liu2023llm, chen2022program, poesia2022synchromesh, nijkamp2022codegen, cai2024lowcode}. However, the generated code may contain errors or fail to meet user requirements. To mitigate these issues, workflow-based methods have been proposed, including manually designed and automatically generated workflows \cite{josifoski2023flows, wu2023autogen, zeng2024flowmind}. Another research direction involves using LLMs for natural language programming, leveraging their strong natural language understanding abilities. A notable example is the CoRE language \cite{xu2024core}, which unifies natural language programming, pseudo-code programming, and workflow programming under the same framework using LLM as interpreter. Our work follows the workflow concept in natural language programming and develops an automated workflow generation framework to reduce human labor. 

\subsection{Automated Machine Learning}

Automated Machine Learning (AutoML) aims to reduce human labors in designing and deploying machine learning techniques, simplifying the application of ML in real-world problems. There are three main types of AutoML techniques \cite{yao2018taking, li2022autolossgen}:

\textit{Automated Model Selection}: Tools such as Auto-sklearn \cite{NIPS2015_11d0e628} and Auto-WEKA \cite{kotthoff2019auto} automatically select the best machine learning model from a library of models and hyper-parameter settings.

\textit{Automated Feature Engineering}: Tools such as Data Science Machine \cite{kanter2015deep}, ExploreKit \cite{katz2016explorekit}, and VEST \cite{cerqueira2021vest} generate or select useful features without manual intervention, since feature engineering significantly impacts model performance in many applications.

\textit{Neural Architecture Search (NAS)}: Methods such as ENAS \cite{pham2018efficient}, DARTS \cite{liu2018darts}, NASH \cite{elsken2017simple}, GNAS \cite{huang2018gnas}, and AmoebaNet-A \cite{real2019regularized} discover effective neural network architectures for specific tasks without manual design. Experiments show that networks generated through NAS can match or even outperform human-designed architectures across various tasks.


AutoML systems typically involve two main components for training: a controller, which is a machine learning model responsible for sampling model selections, and a child model, which comprises the parameters of the machine learning model to be created and used for the task at hand. In our work, we follow this training paradigm, using a workflow generator LLM as the controller, and the generated workflow along with a workflow interpreter LLM as the child model.
More details of the proposed technique are introduced in Section \ref{sec:autoflow}. 

\section{Preliminary and Background}
\label{sec:background}

\subsection{Natural Language Programs as Workflows}
\label{sec:core}

In this section, we introduce how to use natural language programs as a representation of workflows. Specifically, we will use the Code Representation and Execution (CoRE) system \cite{xu2024core} as an example to show how to construct workflows as natural language programs and how the LLM Agent follows the workflow by executing the natural language program.

\subsubsection{CoRE Language Syntax}

The CoRE language defines four components to organize workflows as natural language instructions. 
\begin{itemize}[leftmargin=*]
    \item \textbf{Step Name} is used to uniquely identify each step of the workflow.
    \item \textbf{Step Type} defines the type of instruction for each step. There are three different types of steps:
    \begin{itemize}
        \item Process: The process step transitions to the next specified step after executing the current step.
        \item Decision: Similar to conditional statements (e.g., ``if-else''), the decision step is used for branching the program flow based on evaluated conditions.
        \item Terminal: The terminal step represents the end of the program.
    \end{itemize}
    \item \textbf{Step Instruction} is a natural language instruction to be executed in the step. 
    \item \textbf{Step Connection} points to the next step, which establishs the program execution flow.
\end{itemize}

An example workflow for image-text processing on the OpenAGI benchmark is shown as follows:
\begin{lstlisting}[language=HTML]
Step 1:::Process:::Identify the input data type based on the objective.:::next::Step 2
Step 2:::Process:::Identify the output data type based on the objective.:::next::Step 3
Step 3:::Process:::Select tools in the provided tool list to generate a plan.:::next::Step 4
Step 4:::Decision:::Check whether every tool in the plan is in the provided tool list.:::Yes::Step 5::No::Step 3
Step 5:::Decision:::Check whether the output data type of the previous tool is the input data type
of the next tool.:::Yes::Step 6::No::Step 3
Step 6:::Terminal:::Output the plan by listing the tool names.:::
\end{lstlisting}

In this paper, we use `:::' to delimit the above four components in each step. 

\subsubsection{LLM as Interpreter for Workflow Execution}

To process and execute the workflow in the CoRE language, the system uses an LLM as an interpreter. The LLM interpreter executes instructions step by step. Concretely, the execution of one step can be divided into four procedures in the CoRE system. 

\textbf{\ding{182}} First, the LLM decides which information from memory may be needed to execute the current step and retrieves the relevant information from memory. 
\textbf{\ding{183}} After obtaining the relevant information, the system integrates the information with the instruction of that step into a structured prompt, which the LLM processes to generate a response. 
\textbf{\ding{184}} To extend LLM's capability,
the system may use external tools to analyze the initial response of each step. According to the initial response to the current step, the LLM determines whether external tools are required. If tool usage is confirmed, LLM will decide the tool name and tool arguments, then execute the external tool, and finally incorporate the results into the memory. \textbf{\ding{185}} After the execution of the current step, LLM will decide which is the next step to execute based on the output of the current step.

\subsection{Motivation}
\label{sec:motivation}

The CoRE system enables users to write workflows in natural language, which unifies natural language programming, pseudo-code programming, and workflow programming. Although the entry barrier is lower than coding in programming languages, constructing workflows in natural language still requires much human labor and domain expertise. Inspired by Automated Machine Learning (AutoML) \cite{hutter2019automated}, we would like to automatically learn the best workflow based on the given task and training data. Considering the instructions in CoRE language are written in natural language and LLM has a strong ability of natural language understanding, we also use LLM as the workflow generator. To distinguish with the Interpreter LLM mentioned in \ref{sec:core}, we denote the LLM that learns to generate workflows as the Workflow Generator LLM, and name the LLM that interprets and executes workflow as the Workflow Interpreter LLM, consistent with Figure \ref{fig:workflow}. In this way, users only need to provide a high-level description of the task and the corresponding dataset, and the generator LLM can generate the optimal workflow in CoRE language for the interpreter LLM to execute on the given task. This process expects to minimize human efforts and automatically pursue the optimal workflow for LLM regardless of users' knowledge on workflow design.

\section{The AutoFlow Framework}
\label{sec:autoflow}

\begin{figure}[t]
\hspace{-10pt}
\begin{minipage}{0.5\textwidth}
  \centering
  \includegraphics[width=\linewidth]{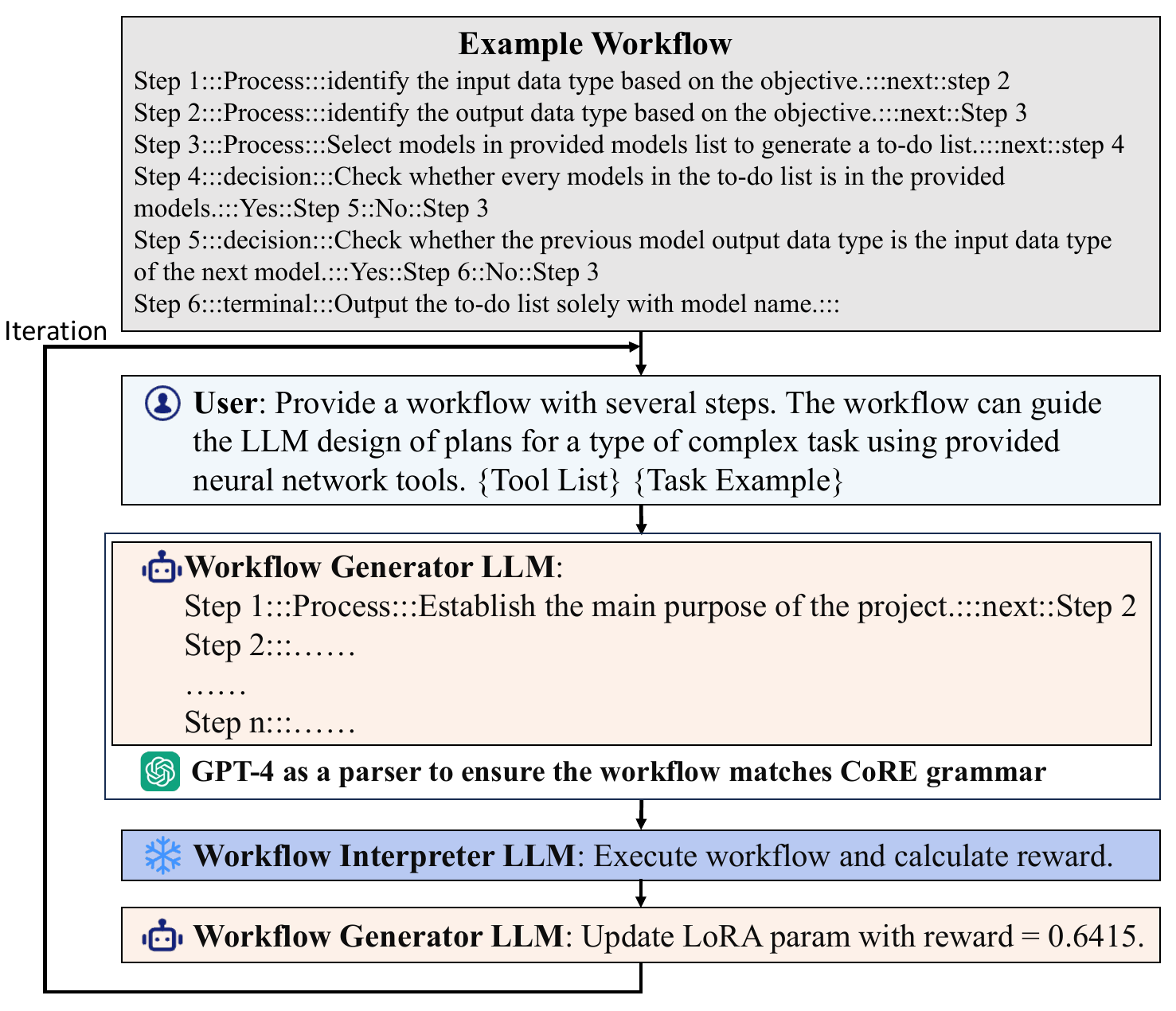}
  \subcaption{AutoFlow generation process based on fine-tuning method with RL reward for open-source LLMs.}
  \label{fig:open_workflow}
\end{minipage}\hfill
\begin{minipage}{0.5\textwidth}
  \centering
  \includegraphics[width=\linewidth]{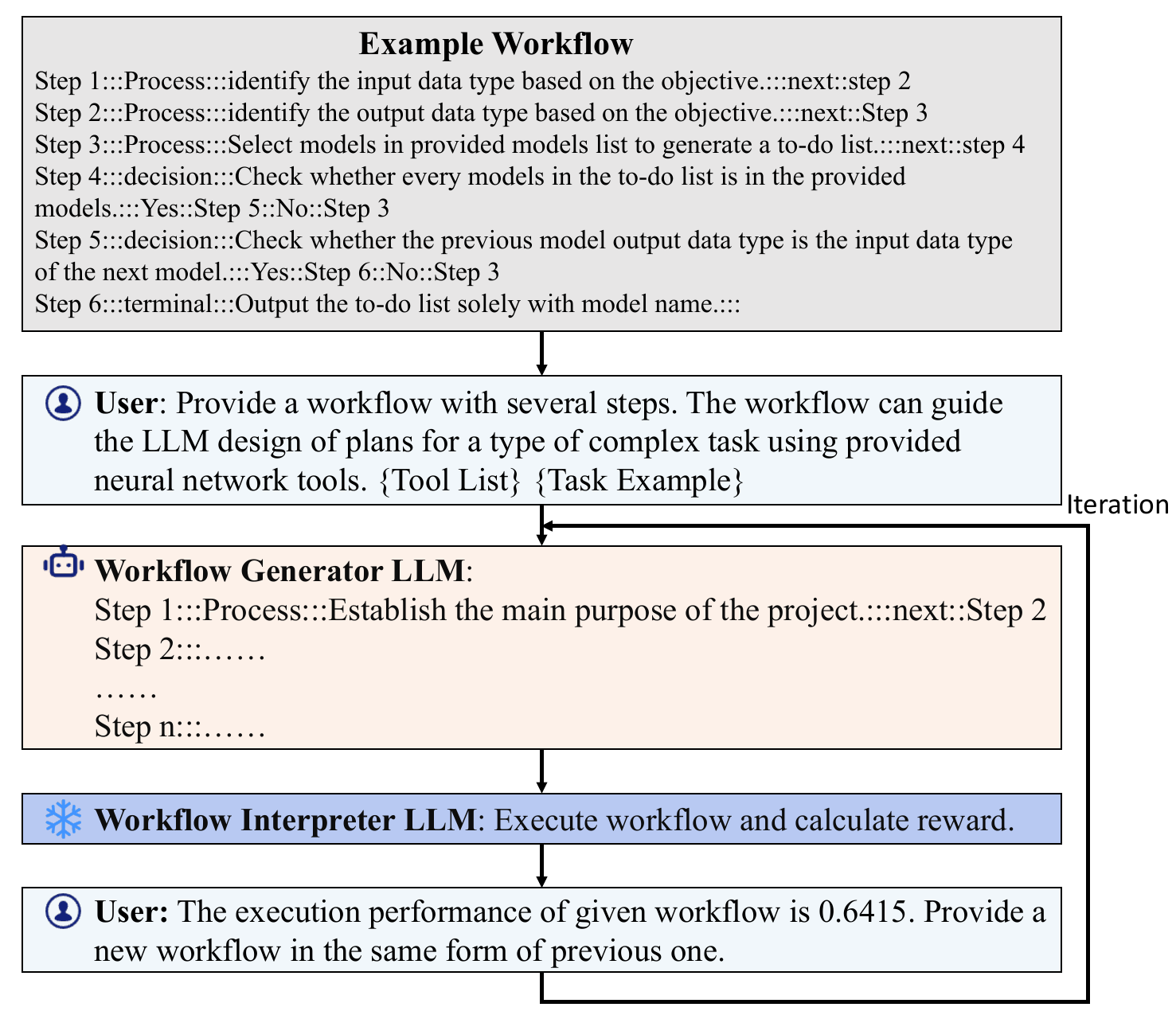}
  \subcaption{AutoFlow generation process based on in-context learning with RL reward for closed-source LLMs.}
  \label{fig:closed_workflow}
\end{minipage}
\caption{Overview for workflow generation with AutoFlow, using OpenAGI \citep{openagi} tasks as an example}
\vspace{-10pt}
\label{fig:overview}
\end{figure}

In this section, we introduce the two methods of applying the AutoFlow framework to the workflow generator LLM, i.e., the fine-tuning method for open-source LLMs and the in-context learning method for closed-source LLMs.

\subsection{Fine-tuning Method for Workflow Generation with Open-source LLMs}

We use LoRA adapter \citep{hu2021lora} for fine-tuning open-souced LLMs 
as workflow generators. The training process is shown in Figure \ref{fig:open_workflow}. 

First, the workflow generator LLM receives a few-shot example workflow and a description of the task from users as the input query. Although the CoRE language has minimal grammar requirements and the instructions are written in natural language, which can be well learned and generated by LLMs, an example workflow can help the workflow generator LLM better understand the grammar of the CoRE language. The natural language description of the task is to help the generator LLM understand the application scenarios of the workflow to be generated. Take the text and image processing tasks in OpenAGI benchmark \cite{openagi} as an example, the task description could be ``Provide a workflow with several steps. The workflow can guide the LLM to design plans for a type of complex tasks realted to text and image processing using the provided tools''. 

Second, the next step is to generate an executable workflow based on the input query. For closed-source LLMs such as GPT-4, the model can directly generate a grammatically valid workflow given the few-shot example. However, open-source LLMs such as Mixtral-8x7B cannot consistently generate grammatically valid workflow even if few-shot example workflows are provided. To solve the problem, we follow the post-processing strategy in previous work \cite{openagi, yuan2024easytool} and use GPT-4 as a parser to revise the output workflow into a grammatically valid one.

Third, the generated workflow will be executed by the interpreter LLM to obtain its performance on the validation dataset. Then, the generator LLM is updated based on the workflow's performance on the validation dataset. 
Specifically, we use reinforcement learning (RL) to update the parameters of the LoRA adapter of the generator LLM, with the average metrics of all data instances on the validation dataset as the reward. 

These three steps together consist of one iteration of the fine-tuning process. The fine-tuning process will terminal and the final workflow will be produced when the terminal condition is met, when the difference of reward between two consecutive iterations is smaller then a threshold. After the iterative optimization process, the workflow generator LLM produces the optimal workflow for the task based on the execution feedback. 

\subsection{In-context Learning Method for Workflow Generation with Closed-source LLMs}

As for closed-source LLMs such as GPT-4, we use in-context learning to avoid fine-tuning the parameters. As shown in Figure \ref{fig:closed_workflow}, the AutoFlow framework also requires an example workflow and a description of the task, and feeds them as the input query to the workflow generator LLM. After the GPT-4 generates the workflow, we do not use a parser to revise the flow since GPT-4 can well follow the CoRE grammar demonstrated by the example workflow. Then, the interpreter LLM executes the workflow to evaluate its performance on the validation dataset as the reward, which is the same process as the fine-tuning method. The difference is that, in the next step, the AutoFlow framework directly includes the reward value in the query and prompts the generator LLM to generate a new workflow given the performance of the previously generated workflow, such as ``The execution performance of the previous workflow is 0.6415. Provide a new workflow that can gain a better performance''. The whole process is demonstrated in Figure \ref{fig:closed_workflow}.



We will show in the experimentation that closed-source LLMs such as GPT-4 can well utilize the reward values in the prompt to refine the workflow and finally obtainn the optimal workflow by using the in-context learning method. 

\section{Experiments}
\label{sec:experiment}

\subsection{Backbone Large Language Model (LLM)}
\label{sec:backbone}

We conduct experiments on both closed-source and open-source LLMs:

\begin{itemize}[itemsep=0pt, topsep=0pt]
    \item \textbf{GPT-4} \citep{openai2023gpt4}  (Closed-source) is a generative pre-trained transformer of OpenAI
    . In this work, we use the GPT-4-1106-preview version.
    \item \textbf{Mixtral-8x7B} \citep{jiang2024mixtral} (Open-source) is a pre-trained generative Sparse Mixture of Experts with 46.7 billion parameters. 
\end{itemize}

In our experiment, we apply these two types of LLMs for both workflow generator LLM and interpreter LLM. Thus, there are four combinations in total.

\subsection{Planning Schema of LLMs}
\label{sec:schema}

We adopt the following LLM-based agent planning schema:

\begin{itemize}[itemsep=0pt, topsep=0pt]
    \item \textbf{Zero-shot Learning (Zero)} directly inputs the query to the LLM.
    \item \textbf{Chain-of-Thought (CoT)} \cite{wei2022chain} induces the LLM to generate a coherent language sequence that serves as a meaningful intermediate step bridging the input query and the output answer.
    \item \textbf{Few-shot Learning (Few)} presents a set of high-quality demonstrations in the prompt, each consisting of both input and desired output on the target task. 
    \item \textbf{CoRE} \cite{xu2024core} uses a manually designed workflow with LLM as an interpreter.
    \item \textbf{AutoFlow} is our proposed framework that can automatically generate workflows.
\end{itemize}

\subsection{Benchmark Datasets}
\label{sec:dataset}

We conduct experiments on a benchmark dataset, \textbf{OpenAGI} \citep{openagi}. 
The OpenAGI benchmark tasks are categorized based on their output type and ground-truth label type (\textbf{Task 1}, \textbf{2}, and \textbf{3}). Then, based on different task types, different metrics are employed to gauge the performance: \textbf{CLIP Score} \citep{hessel2021clipscore}, assessing the similarity between text and image, is utilized for Text-to-Image tasks (Task 1); \textbf{BERT Score} \citep{bert-score}, evaluating text generation with BERT, is applied when both data labels and the expected outputs are texts (Task 2); and \textbf{ViT Score} \citep{wu2020visual} gauges the similarity between the image label and image output (Task 3). 

\subsection{Implementation Details}
\label{sec:implementation}

Our framework and all baselines are implemented by PyTorch, an open-source library. We follow the implementation setting of the OpenAGI platform \citep{openagi} for Zero-shot and few-shot learnings. We leverage the DSPy framework \cite{khattab2022demonstrate, khattab2023dspy} to apply the CoT strategy to the OpenAGI platform. We also tried Program-of-Thought \cite{chen2022program} and ReAct \cite{yao2023react} strategies on the OpenAGI platform. However, the ReAct strategy requires text observation, which is unsuitable for our OpenAGI task since some observations are in image format, and Program-of-Thought cannot generate executable codes. Thus, we did not include them as the baselines.

For the hyper-parameter setting of the AutoFlow framework, we set the number of iterations for the workflow generator LLM as 30. For the open-source LLM, Mixtral, as the generator LLM, we use the REINFORCE \citep{williams1992simple} as the core reinforcement learning (RL) algorithm for the generator LLM, with the average score on the training dataset as the reward. We use Adam as the optimizer with the learning rate at 0.001 for RL. Also, we apply Low-Rank Adaptation (LoRA) \citep{hu2021lora} with the rank equal to 8 to Mixtral for efficient fine-tuning.

\subsection{Experimental Analysis}
\label{sec:exp_analysis}

We conduct the experiments on the OpenAGI \citep{openagi} benchmark dataset. For a fair comparison, we show the results using the same workflow interpreter LLM in a table. Specifically, the results of using the open-source LLM, Mixtral, as the LLM interpreter is shown in Table \ref{Table:open_source}; and the results of using the closed-source LLM, GPT-4, as the LLM interpreter is shown in Table \ref{Table:closed_source}. Each row stands for a type of task, each column represents the planning schema of an LLM interpreter. From these two tables, we can see that, after applying our AutoFlow framework, the average score over tasks is significantly better than the baselines. Compared to the best baseline, CoRE, AutoFlow has over 40\% improvement when using Mixtral as the LLM interpreter, and over 5\% improvement when using GPT-4 as the interpreter LLM. For the score of each type of task, our AutoFlow also reaches the highest one. Thus, the experiment results validate that AutoFlow is effective and can generate a workflow with better performance than manually designed ones.

An interesting observation is that, the best average score when using Mixtral as the LLM interpreter, is AutoFlow with GPT-4 as the workflow generator; and the best average score when using GPT-4 as the LLM interpreter, is AutoFlow with Mixtral as the workflow generator. This observation suggests that the combination of different systems (Mixtral and GPT-4) for the LLM interpreter and workflow generator might lead to a kind of synergistic effect where the strengths of one system complement the weaknesses of the other,
which helps to better solve complex multi-step tasks.

\begin{table}[t]
\small
    \centering
    \begin{tabular}{c|cccc|cc}
    \hline
      Metrics / Task & Zero & CoT & Few & CoRE & AutoFlow (GPT) & AutoFlow (Mixtral) \\
      \hline
      Task 1 (CLIP Score) & 0.0 & 0.0 & 0.1839 & 0.1825 & \bm{$0.2441$} & 0.1831\\
      Task 2 (BERT Score) & 0.1092 & 0.1987 & 0.0687 & 0.2593 & 0.3017 & \bm{$0.3133$} \\
      Task 3 (ViT Score) & 0.1949 & 0.1562 & 0.5501 & 0.2437 & \bm{$0.5720$} & 0.4907 \\
      Average over tasks & 0.1206 & 0.1736 & 0.1887 & 0.2483 & \bm{$0.3597$} & 0.3442 \\
      \hline
    \end{tabular}
    \caption{Performance on OpenAGI when using the open-source LLM, Mixtral, as the LLM interpreter for all tasks and learning schema. Zero is for Zero-shot Learning, Few is for Few-shot Learning. The boldface numbers denote the highest score under each task type using the same LLM.}
    \label{Table:open_source}
    \vspace{-15pt}
\end{table}

\begin{table}[t]
\small
    \centering
    \begin{tabular}{c|cccc|cc}
    \hline
      Metrics / Task & Zero & CoT & Few & CoRE & AutoFlow (GPT) & AutoFlow (Mixtral) \\
      \hline
      Task 1 (CLIP Score) & 0.0 & 0.2732 & 0.3055 & 0.1368 & \bm{$0.3049$} & 0.3032 \\
      Task 2 (BERT Score) & 0.2076 & 0.2266 & 0.6307 & 0.6505 & 0.6628 & \bm{$0.7014$}\\
      Task 3 (ViT Score) & 0.5058 & 0.6736 & 0.6480 & 0.6480 & \bm{$0.6899$} & 0.6119 \\
      Average over tasks & 0.2378 & 0.3359 & 0.5281 & 0.6104 & 0.6415 & \bm{$0.6501$} \\
      \hline
    \end{tabular}
    \caption{Performance on OpenAGI using the closed-source LLM, GPT-4, as the LLM interpreter for all tasks and learning schema. Zero is for Zero-shot Learning, Few is for Few-shot Learning. The boldface numbers denote the highest score under each task type using the same LLM.}
    \label{Table:closed_source}
    \vspace{-15pt}
\end{table}

\section{Conclusions and Future Work}
\label{sec:conclusions}

In this study, we introduce the AutoFlow framework to use Large Language Models (LLMs) for automatically generating effective workflows for agents. We propose two learning methods for AutoFlow, the fine-tuning method when using open-source LLM as workflow generator, and the in-context learning method when using closed-source LLM as workflow generator. 
Compared to manually designed workflows, automatically generated workflows can reach better performance and significantly reduce the human labor, leading to a higher degree of automation. 

Although AutoFlow demonstrates promising results, there is still space for improvement. For example, the learning process for the workflow generator LLM uses reinforcement learning, which may not be the most efficient compared to some gradient-based methods or few-shot learning methods. Future studies may try to evaluate the efficacy of other learning methods. Another example in the AutoFlow framework is that, the workflow generator and interpreter LLMs work together using a collaborative learning paradigm. Instead, we may try  other learning paradigms such as the teacher-student paradigm or the adversarial learning paradigm.

\bibliographystyle{ACM-Reference-Format}
\bibliography{reference}


\begin{thebibliography}{54}


\ifx \showCODEN    \undefined \def \showCODEN     #1{\unskip}     \fi
\ifx \showDOI      \undefined \def \showDOI       #1{#1}\fi
\ifx \showISBNx    \undefined \def \showISBNx     #1{\unskip}     \fi
\ifx \showISBNxiii \undefined \def \showISBNxiii  #1{\unskip}     \fi
\ifx \showISSN     \undefined \def \showISSN      #1{\unskip}     \fi
\ifx \showLCCN     \undefined \def \showLCCN      #1{\unskip}     \fi
\ifx \shownote     \undefined \def \shownote      #1{#1}          \fi
\ifx \showarticletitle \undefined \def \showarticletitle #1{#1}   \fi
\ifx \showURL      \undefined \def \showURL       {\relax}        \fi
\providecommand\bibfield[2]{#2}
\providecommand\bibinfo[2]{#2}
\providecommand\natexlab[1]{#1}
\providecommand\showeprint[2][]{arXiv:#2}

\bibitem[Besta et~al\mbox{.}(2024)]%
        {besta2024graph}
\bibfield{author}{\bibinfo{person}{Maciej Besta}, \bibinfo{person}{Nils Blach}, \bibinfo{person}{Ales Kubicek}, \bibinfo{person}{Robert Gerstenberger}, \bibinfo{person}{Michal Podstawski}, \bibinfo{person}{Lukas Gianinazzi}, \bibinfo{person}{Joanna Gajda}, \bibinfo{person}{Tomasz Lehmann}, \bibinfo{person}{Hubert Niewiadomski}, \bibinfo{person}{Piotr Nyczyk}, {et~al\mbox{.}}} \bibinfo{year}{2024}\natexlab{}.
\newblock \showarticletitle{Graph of thoughts: Solving elaborate problems with large language models}. In \bibinfo{booktitle}{\emph{Proceedings of the AAAI Conference on Artificial Intelligence}}, Vol.~\bibinfo{volume}{38}. \bibinfo{pages}{17682--17690}.
\newblock


\bibitem[Boiko et~al\mbox{.}(2023)]%
        {boiko2023emergent}
\bibfield{author}{\bibinfo{person}{Daniil~A. Boiko}, \bibinfo{person}{Robert MacKnight}, {and} \bibinfo{person}{Gabe Gomes}.} \bibinfo{year}{2023}\natexlab{}.
\newblock \bibinfo{title}{Emergent autonomous scientific research capabilities of large language models}.
\newblock
\newblock
\showeprint[arxiv]{2304.05332}~[physics.chem-ph]


\bibitem[Cai et~al\mbox{.}(2024)]%
        {cai2024lowcode}
\bibfield{author}{\bibinfo{person}{Yuzhe Cai}, \bibinfo{person}{Shaoguang Mao}, \bibinfo{person}{Wenshan Wu}, \bibinfo{person}{Zehua Wang}, \bibinfo{person}{Yaobo Liang}, \bibinfo{person}{Tao Ge}, \bibinfo{person}{Chenfei Wu}, \bibinfo{person}{Wang You}, \bibinfo{person}{Ting Song}, \bibinfo{person}{Yan Xia}, \bibinfo{person}{Jonathan Tien}, \bibinfo{person}{Nan Duan}, {and} \bibinfo{person}{Furu Wei}.} \bibinfo{year}{2024}\natexlab{}.
\newblock \bibinfo{title}{Low-code LLM: Graphical User Interface over Large Language Models}.
\newblock
\newblock
\showeprint[arxiv]{2304.08103}~[cs.CL]


\bibitem[Cerqueira et~al\mbox{.}(2021)]%
        {cerqueira2021vest}
\bibfield{author}{\bibinfo{person}{Vitor Cerqueira}, \bibinfo{person}{Nuno Moniz}, {and} \bibinfo{person}{Carlos Soares}.} \bibinfo{year}{2021}\natexlab{}.
\newblock \showarticletitle{Vest: Automatic feature engineering for forecasting}.
\newblock \bibinfo{journal}{\emph{Machine Learning}} (\bibinfo{year}{2021}), \bibinfo{pages}{1--23}.
\newblock


\bibitem[Chen et~al\mbox{.}(2023)]%
        {chen2022program}
\bibfield{author}{\bibinfo{person}{Wenhu Chen}, \bibinfo{person}{Xueguang Ma}, \bibinfo{person}{Xinyi Wang}, {and} \bibinfo{person}{William~W. Cohen}.} \bibinfo{year}{2023}\natexlab{}.
\newblock \showarticletitle{Program of Thoughts Prompting: Disentangling Computation from Reasoning for Numerical Reasoning Tasks}.
\newblock \bibinfo{journal}{\emph{Transactions on Machine Learning Research}} (\bibinfo{year}{2023}).
\newblock


\bibitem[Feurer et~al\mbox{.}(2015)]%
        {NIPS2015_11d0e628}
\bibfield{author}{\bibinfo{person}{Matthias Feurer}, \bibinfo{person}{Aaron Klein}, \bibinfo{person}{Katharina Eggensperger}, \bibinfo{person}{Jost Springenberg}, \bibinfo{person}{Manuel Blum}, {and} \bibinfo{person}{Frank Hutter}.} \bibinfo{year}{2015}\natexlab{}.
\newblock \showarticletitle{Efficient and Robust Automated Machine Learning}. In \bibinfo{booktitle}{\emph{Advances in Neural Information Processing Systems}}, \bibfield{editor}{\bibinfo{person}{C.~Cortes}, \bibinfo{person}{N.~Lawrence}, \bibinfo{person}{D.~Lee}, \bibinfo{person}{M.~Sugiyama}, {and} \bibinfo{person}{R.~Garnett}} (Eds.), Vol.~\bibinfo{volume}{28}. \bibinfo{publisher}{Curran Associates, Inc.}
\newblock
\urldef\tempurl%
\url{https://proceedings.neurips.cc/paper/2015/file/11d0e6287202fced83f79975ec59a3a6-Paper.pdf}
\showURL{%
\tempurl}


\bibitem[Ge et~al\mbox{.}(2023a)]%
        {openagi}
\bibfield{author}{\bibinfo{person}{Yingqiang Ge}, \bibinfo{person}{Wenyue Hua}, \bibinfo{person}{Kai Mei}, \bibinfo{person}{Jianchao Ji}, \bibinfo{person}{Juntao Tan}, \bibinfo{person}{Shuyuan Xu}, \bibinfo{person}{Zelong Li}, {and} \bibinfo{person}{Yongfeng Zhang}.} \bibinfo{year}{2023}\natexlab{a}.
\newblock \showarticletitle{OpenAGI: When LLM Meets Domain Experts}.
\newblock \bibinfo{journal}{\emph{In Advances in Neural Information Processing Systems (NeurIPS)}} (\bibinfo{year}{2023}).
\newblock


\bibitem[Ge et~al\mbox{.}(2023b)]%
        {ge2023llm}
\bibfield{author}{\bibinfo{person}{Yingqiang Ge}, \bibinfo{person}{Yujie Ren}, \bibinfo{person}{Wenyue Hua}, \bibinfo{person}{Shuyuan Xu}, \bibinfo{person}{Juntao Tan}, {and} \bibinfo{person}{Yongfeng Zhang}.} \bibinfo{year}{2023}\natexlab{b}.
\newblock \showarticletitle{LLM as OS, Agents as Apps: Envisioning AIOS, Agents and the AIOS-Agent Ecosystem}.
\newblock \bibinfo{journal}{\emph{arXiv e-prints}} (\bibinfo{year}{2023}), \bibinfo{pages}{arXiv--2312}.
\newblock


\bibitem[Hao et~al\mbox{.}(2023)]%
        {hao2023reasoning}
\bibfield{author}{\bibinfo{person}{Shibo Hao}, \bibinfo{person}{Yi Gu}, \bibinfo{person}{Haodi Ma}, \bibinfo{person}{Joshua~Jiahua Hong}, \bibinfo{person}{Zhen Wang}, \bibinfo{person}{Daisy~Zhe Wang}, {and} \bibinfo{person}{Zhiting Hu}.} \bibinfo{year}{2023}\natexlab{}.
\newblock \showarticletitle{Reasoning with language model is planning with world model}.
\newblock \bibinfo{journal}{\emph{arXiv preprint arXiv:2305.14992}} (\bibinfo{year}{2023}).
\newblock


\bibitem[Hessel et~al\mbox{.}(2021)]%
        {hessel2021clipscore}
\bibfield{author}{\bibinfo{person}{Jack Hessel}, \bibinfo{person}{Ari Holtzman}, \bibinfo{person}{Maxwell Forbes}, \bibinfo{person}{Ronan~Le Bras}, {and} \bibinfo{person}{Yejin Choi}.} \bibinfo{year}{2021}\natexlab{}.
\newblock \bibinfo{title}{{CLIPScore:} A Reference-free Evaluation Metric for Image Captioning}.
\newblock
\newblock


\bibitem[Hu et~al\mbox{.}(2021)]%
        {hu2021lora}
\bibfield{author}{\bibinfo{person}{Edward~J. Hu}, \bibinfo{person}{Yelong Shen}, \bibinfo{person}{Phillip Wallis}, \bibinfo{person}{Zeyuan Allen-Zhu}, \bibinfo{person}{Yuanzhi Li}, \bibinfo{person}{Shean Wang}, \bibinfo{person}{Lu Wang}, {and} \bibinfo{person}{Weizhu Chen}.} \bibinfo{year}{2021}\natexlab{}.
\newblock \bibinfo{title}{LoRA: Low-Rank Adaptation of Large Language Models}.
\newblock
\newblock
\showeprint[arxiv]{2106.09685}~[cs.CL]


\bibitem[Huang et~al\mbox{.}(2018)]%
        {huang2018gnas}
\bibfield{author}{\bibinfo{person}{Siyu Huang}, \bibinfo{person}{Xi Li}, \bibinfo{person}{Zhi-Qi Cheng}, \bibinfo{person}{Zhongfei Zhang}, {and} \bibinfo{person}{Alexander Hauptmann}.} \bibinfo{year}{2018}\natexlab{}.
\newblock \showarticletitle{Gnas: A greedy neural architecture search method for multi-attribute learning}. In \bibinfo{booktitle}{\emph{Proceedings of the 26th ACM international conference on Multimedia}}. \bibinfo{pages}{2049--2057}.
\newblock


\bibitem[Hutter et~al\mbox{.}(2019)]%
        {hutter2019automated}
\bibfield{author}{\bibinfo{person}{Frank Hutter}, \bibinfo{person}{Lars Kotthoff}, {and} \bibinfo{person}{Joaquin Vanschoren}.} \bibinfo{year}{2019}\natexlab{}.
\newblock \bibinfo{booktitle}{\emph{Automated machine learning: methods, systems, challenges}}.
\newblock \bibinfo{publisher}{Springer Nature}.
\newblock


\bibitem[Jiang et~al\mbox{.}(2024)]%
        {jiang2024mixtral}
\bibfield{author}{\bibinfo{person}{Albert~Q Jiang}, \bibinfo{person}{Alexandre Sablayrolles}, \bibinfo{person}{Antoine Roux}, \bibinfo{person}{Arthur Mensch}, \bibinfo{person}{Blanche Savary}, \bibinfo{person}{Chris Bamford}, \bibinfo{person}{Devendra~Singh Chaplot}, \bibinfo{person}{Diego de~las Casas}, \bibinfo{person}{Emma~Bou Hanna}, \bibinfo{person}{Florian Bressand}, {et~al\mbox{.}}} \bibinfo{year}{2024}\natexlab{}.
\newblock \showarticletitle{Mixtral of experts}.
\newblock \bibinfo{journal}{\emph{arXiv preprint arXiv:2401.04088}} (\bibinfo{year}{2024}).
\newblock


\bibitem[Jojic et~al\mbox{.}(2023)]%
        {jojic2023gpt}
\bibfield{author}{\bibinfo{person}{Ana Jojic}, \bibinfo{person}{Zhen Wang}, {and} \bibinfo{person}{Nebojsa Jojic}.} \bibinfo{year}{2023}\natexlab{}.
\newblock \showarticletitle{Gpt is becoming a turing machine: Here are some ways to program it}.
\newblock \bibinfo{journal}{\emph{arXiv preprint arXiv:2303.14310}} (\bibinfo{year}{2023}).
\newblock


\bibitem[Josifoski et~al\mbox{.}(2023)]%
        {josifoski2023flows}
\bibfield{author}{\bibinfo{person}{Martin Josifoski}, \bibinfo{person}{Lars Klein}, \bibinfo{person}{Maxime Peyrard}, \bibinfo{person}{Yifei Li}, \bibinfo{person}{Saibo Geng}, \bibinfo{person}{Julian~Paul Schnitzler}, \bibinfo{person}{Yuxing Yao}, \bibinfo{person}{Jiheng Wei}, \bibinfo{person}{Debjit Paul}, {and} \bibinfo{person}{Robert West}.} \bibinfo{year}{2023}\natexlab{}.
\newblock \bibinfo{title}{Flows: Building Blocks of Reasoning and Collaborating AI}.
\newblock
\newblock
\showeprint[arxiv]{2308.01285}~[cs.AI]


\bibitem[Kanter and Veeramachaneni(2015)]%
        {kanter2015deep}
\bibfield{author}{\bibinfo{person}{James~Max Kanter} {and} \bibinfo{person}{Kalyan Veeramachaneni}.} \bibinfo{year}{2015}\natexlab{}.
\newblock \showarticletitle{Deep feature synthesis: Towards automating data science endeavors}. In \bibinfo{booktitle}{\emph{2015 IEEE international conference on data science and advanced analytics (DSAA)}}. IEEE, \bibinfo{pages}{1--10}.
\newblock


\bibitem[Katz et~al\mbox{.}(2016)]%
        {katz2016explorekit}
\bibfield{author}{\bibinfo{person}{Gilad Katz}, \bibinfo{person}{Eui Chul~Richard Shin}, {and} \bibinfo{person}{Dawn Song}.} \bibinfo{year}{2016}\natexlab{}.
\newblock \showarticletitle{Explorekit: Automatic feature generation and selection}. In \bibinfo{booktitle}{\emph{2016 IEEE 16th International Conference on Data Mining (ICDM)}}. IEEE, \bibinfo{pages}{979--984}.
\newblock


\bibitem[Khattab et~al\mbox{.}(2022)]%
        {khattab2022demonstrate}
\bibfield{author}{\bibinfo{person}{Omar Khattab}, \bibinfo{person}{Keshav Santhanam}, \bibinfo{person}{Xiang~Lisa Li}, \bibinfo{person}{David Hall}, \bibinfo{person}{Percy Liang}, \bibinfo{person}{Christopher Potts}, {and} \bibinfo{person}{Matei Zaharia}.} \bibinfo{year}{2022}\natexlab{}.
\newblock \showarticletitle{Demonstrate-Search-Predict: Composing Retrieval and Language Models for Knowledge-Intensive {NLP}}.
\newblock \bibinfo{journal}{\emph{arXiv preprint arXiv:2212.14024}} (\bibinfo{year}{2022}).
\newblock


\bibitem[Khattab et~al\mbox{.}(2023)]%
        {khattab2023dspy}
\bibfield{author}{\bibinfo{person}{Omar Khattab}, \bibinfo{person}{Arnav Singhvi}, \bibinfo{person}{Paridhi Maheshwari}, \bibinfo{person}{Zhiyuan Zhang}, \bibinfo{person}{Keshav Santhanam}, \bibinfo{person}{Sri Vardhamanan}, \bibinfo{person}{Saiful Haq}, \bibinfo{person}{Ashutosh Sharma}, \bibinfo{person}{Thomas~T. Joshi}, \bibinfo{person}{Hanna Moazam}, \bibinfo{person}{Heather Miller}, \bibinfo{person}{Matei Zaharia}, {and} \bibinfo{person}{Christopher Potts}.} \bibinfo{year}{2023}\natexlab{}.
\newblock \showarticletitle{DSPy: Compiling Declarative Language Model Calls into Self-Improving Pipelines}.
\newblock \bibinfo{journal}{\emph{arXiv preprint arXiv:2310.03714}} (\bibinfo{year}{2023}).
\newblock


\bibitem[Kojima et~al\mbox{.}(2022)]%
        {kojima2022large}
\bibfield{author}{\bibinfo{person}{Takeshi Kojima}, \bibinfo{person}{Shixiang~Shane Gu}, \bibinfo{person}{Machel Reid}, \bibinfo{person}{Yutaka Matsuo}, {and} \bibinfo{person}{Yusuke Iwasawa}.} \bibinfo{year}{2022}\natexlab{}.
\newblock \showarticletitle{Large language models are zero-shot reasoners}.
\newblock \bibinfo{journal}{\emph{Advances in neural information processing systems}}  \bibinfo{volume}{35} (\bibinfo{year}{2022}), \bibinfo{pages}{22199--22213}.
\newblock


\bibitem[Kotthoff et~al\mbox{.}(2019)]%
        {kotthoff2019auto}
\bibfield{author}{\bibinfo{person}{Lars Kotthoff}, \bibinfo{person}{Chris Thornton}, \bibinfo{person}{Holger~H Hoos}, \bibinfo{person}{Frank Hutter}, {and} \bibinfo{person}{Kevin Leyton-Brown}.} \bibinfo{year}{2019}\natexlab{}.
\newblock \showarticletitle{Auto-WEKA: Automatic model selection and hyperparameter optimization in WEKA}.
\newblock In \bibinfo{booktitle}{\emph{Automated Machine Learning}}. \bibinfo{publisher}{Springer, Cham}, \bibinfo{pages}{81--95}.
\newblock


\bibitem[Li et~al\mbox{.}(2023)]%
        {li2023camel}
\bibfield{author}{\bibinfo{person}{Guohao Li}, \bibinfo{person}{Hasan Abed Al~Kader Hammoud}, \bibinfo{person}{Hani Itani}, \bibinfo{person}{Dmitrii Khizbullin}, {and} \bibinfo{person}{Bernard Ghanem}.} \bibinfo{year}{2023}\natexlab{}.
\newblock \showarticletitle{CAMEL: Communicative Agents for "Mind" Exploration of Large Language Model Society}. In \bibinfo{booktitle}{\emph{Thirty-seventh Conference on Neural Information Processing Systems}}.
\newblock


\bibitem[Li et~al\mbox{.}(2024b)]%
        {li2024large}
\bibfield{author}{\bibinfo{person}{Xinyi Li}, \bibinfo{person}{Yongfeng Zhang}, {and} \bibinfo{person}{Edward~C Malthouse}.} \bibinfo{year}{2024}\natexlab{b}.
\newblock \showarticletitle{Large Language Model Agent for Fake News Detection}.
\newblock \bibinfo{journal}{\emph{arXiv preprint arXiv:2405.01593}} (\bibinfo{year}{2024}).
\newblock


\bibitem[Li et~al\mbox{.}(2024a)]%
        {li2024formalllm}
\bibfield{author}{\bibinfo{person}{Zelong Li}, \bibinfo{person}{Wenyue Hua}, \bibinfo{person}{Hao Wang}, \bibinfo{person}{He Zhu}, {and} \bibinfo{person}{Yongfeng Zhang}.} \bibinfo{year}{2024}\natexlab{a}.
\newblock \showarticletitle{Formal-LLM: Integrating Formal Language and Natural Language for Controllable LLM-based Agents}.
\newblock \bibinfo{journal}{\emph{arXiv:2402.00798}} (\bibinfo{year}{2024}).
\newblock


\bibitem[Li et~al\mbox{.}(2022)]%
        {li2022autolossgen}
\bibfield{author}{\bibinfo{person}{Zelong Li}, \bibinfo{person}{Jianchao Ji}, \bibinfo{person}{Yingqiang Ge}, {and} \bibinfo{person}{Yongfeng Zhang}.} \bibinfo{year}{2022}\natexlab{}.
\newblock \showarticletitle{AutoLossGen: Automatic Loss Function Generation for Recommender Systems}.
\newblock \bibinfo{journal}{\emph{SIGIR}} (\bibinfo{year}{2022}).
\newblock


\bibitem[Liu et~al\mbox{.}(2023)]%
        {liu2023llm}
\bibfield{author}{\bibinfo{person}{Bo Liu}, \bibinfo{person}{Yuqian Jiang}, \bibinfo{person}{Xiaohan Zhang}, \bibinfo{person}{Qiang Liu}, \bibinfo{person}{Shiqi Zhang}, \bibinfo{person}{Joydeep Biswas}, {and} \bibinfo{person}{Peter Stone}.} \bibinfo{year}{2023}\natexlab{}.
\newblock \showarticletitle{Llm+ p: Empowering large language models with optimal planning proficiency}.
\newblock \bibinfo{journal}{\emph{arXiv preprint arXiv:2304.11477}} (\bibinfo{year}{2023}).
\newblock


\bibitem[Liu et~al\mbox{.}(2019)]%
        {liu2018darts}
\bibfield{author}{\bibinfo{person}{Hanxiao Liu}, \bibinfo{person}{Karen Simonyan}, {and} \bibinfo{person}{Yiming Yang}.} \bibinfo{year}{2019}\natexlab{}.
\newblock \showarticletitle{{DARTS}: Differentiable Architecture Search}. In \bibinfo{booktitle}{\emph{International Conference on Learning Representations}}.
\newblock
\urldef\tempurl%
\url{https://openreview.net/forum?id=S1eYHoC5FX}
\showURL{%
\tempurl}


\bibitem[Lyu et~al\mbox{.}(2023)]%
        {lyu2023faithful}
\bibfield{author}{\bibinfo{person}{Qing Lyu}, \bibinfo{person}{Shreya Havaldar}, \bibinfo{person}{Adam Stein}, \bibinfo{person}{Li Zhang}, \bibinfo{person}{Delip Rao}, \bibinfo{person}{Eric Wong}, \bibinfo{person}{Marianna Apidianaki}, {and} \bibinfo{person}{Chris Callison-Burch}.} \bibinfo{year}{2023}\natexlab{}.
\newblock \showarticletitle{Faithful chain-of-thought reasoning}.
\newblock \bibinfo{journal}{\emph{arXiv preprint arXiv:2301.13379}} (\bibinfo{year}{2023}).
\newblock


\bibitem[Mei et~al\mbox{.}(2024)]%
        {mei2024aios}
\bibfield{author}{\bibinfo{person}{Kai Mei}, \bibinfo{person}{Zelong Li}, \bibinfo{person}{Shuyuan Xu}, \bibinfo{person}{Ruosong Ye}, \bibinfo{person}{Yingqiang Ge}, {and} \bibinfo{person}{Yongfeng Zhang}.} \bibinfo{year}{2024}\natexlab{}.
\newblock \showarticletitle{AIOS: LLM Agent Operating System}.
\newblock \bibinfo{journal}{\emph{arXiv}} (\bibinfo{year}{2024}).
\newblock


\bibitem[Nijkamp et~al\mbox{.}(2022)]%
        {nijkamp2022codegen}
\bibfield{author}{\bibinfo{person}{Erik Nijkamp}, \bibinfo{person}{Bo Pang}, \bibinfo{person}{Hiroaki Hayashi}, \bibinfo{person}{Lifu Tu}, \bibinfo{person}{Huan Wang}, \bibinfo{person}{Yingbo Zhou}, \bibinfo{person}{Silvio Savarese}, {and} \bibinfo{person}{Caiming Xiong}.} \bibinfo{year}{2022}\natexlab{}.
\newblock \showarticletitle{Codegen: An open large language model for code with multi-turn program synthesis}.
\newblock \bibinfo{journal}{\emph{arXiv preprint arXiv:2203.13474}} (\bibinfo{year}{2022}).
\newblock


\bibitem[OpenAI(2023)]%
        {openai2023gpt4}
\bibfield{author}{\bibinfo{person}{Josh et~al OpenAI}.} \bibinfo{year}{2023}\natexlab{}.
\newblock \bibinfo{title}{GPT-4 Technical Report}.
\newblock
\newblock
\showeprint[arxiv]{2303.08774}~[cs.CL]


\bibitem[Pham et~al\mbox{.}(2018)]%
        {pham2018efficient}
\bibfield{author}{\bibinfo{person}{Hieu Pham}, \bibinfo{person}{Melody Guan}, \bibinfo{person}{Barret Zoph}, \bibinfo{person}{Quoc Le}, {and} \bibinfo{person}{Jeff Dean}.} \bibinfo{year}{2018}\natexlab{}.
\newblock \showarticletitle{Efficient neural architecture search via parameters sharing}. In \bibinfo{booktitle}{\emph{International Conference on Machine Learning}}. PMLR, \bibinfo{pages}{4095--4104}.
\newblock


\bibitem[Poesia et~al\mbox{.}(2022)]%
        {poesia2022synchromesh}
\bibfield{author}{\bibinfo{person}{Gabriel Poesia}, \bibinfo{person}{Oleksandr Polozov}, \bibinfo{person}{Vu Le}, \bibinfo{person}{Ashish Tiwari}, \bibinfo{person}{Gustavo Soares}, \bibinfo{person}{Christopher Meek}, {and} \bibinfo{person}{Sumit Gulwani}.} \bibinfo{year}{2022}\natexlab{}.
\newblock \showarticletitle{Synchromesh: Reliable code generation from pre-trained language models}.
\newblock \bibinfo{journal}{\emph{arXiv preprint arXiv:2201.11227}} (\bibinfo{year}{2022}).
\newblock


\bibitem[Qian et~al\mbox{.}(2023)]%
        {qian2023communicative}
\bibfield{author}{\bibinfo{person}{Chen Qian}, \bibinfo{person}{Xin Cong}, \bibinfo{person}{Wei Liu}, \bibinfo{person}{Cheng Yang}, \bibinfo{person}{Weize Chen}, \bibinfo{person}{Yusheng Su}, \bibinfo{person}{Yufan Dang}, \bibinfo{person}{Jiahao Li}, \bibinfo{person}{Juyuan Xu}, \bibinfo{person}{Dahai Li}, \bibinfo{person}{Zhiyuan Liu}, {and} \bibinfo{person}{Maosong Sun}.} \bibinfo{year}{2023}\natexlab{}.
\newblock \bibinfo{title}{Communicative Agents for Software Development}.
\newblock
\newblock
\showeprint[arxiv]{2307.07924}~[cs.SE]


\bibitem[Real et~al\mbox{.}(2019)]%
        {real2019regularized}
\bibfield{author}{\bibinfo{person}{Esteban Real}, \bibinfo{person}{Alok Aggarwal}, \bibinfo{person}{Yanping Huang}, {and} \bibinfo{person}{Quoc~V. Le}.} \bibinfo{year}{2019}\natexlab{}.
\newblock \showarticletitle{Regularized Evolution for Image Classifier Architecture Search}.
\newblock \bibinfo{journal}{\emph{Proceedings of the AAAI Conference on Artificial Intelligence}} \bibinfo{volume}{33}, \bibinfo{number}{01} (\bibinfo{date}{Jul.} \bibinfo{year}{2019}), \bibinfo{pages}{4780--4789}.
\newblock
\urldef\tempurl%
\url{https://doi.org/10.1609/aaai.v33i01.33014780}
\showDOI{\tempurl}


\bibitem[Thomas~Elsken(2018)]%
        {elsken2017simple}
\bibfield{author}{\bibinfo{person}{Frank~Hutter Thomas~Elsken, Jan Hendrik~Metzen}.} \bibinfo{year}{2018}\natexlab{}.
\newblock \bibinfo{title}{Simple and efficient architecture search for Convolutional Neural Networks}.
\newblock
\newblock
\urldef\tempurl%
\url{https://openreview.net/forum?id=SySaJ0xCZ}
\showURL{%
\tempurl}


\bibitem[Wang et~al\mbox{.}(2023)]%
        {wang2023survey}
\bibfield{author}{\bibinfo{person}{Lei Wang}, \bibinfo{person}{Chen Ma}, \bibinfo{person}{Xueyang Feng}, \bibinfo{person}{Zeyu Zhang}, \bibinfo{person}{Hao Yang}, \bibinfo{person}{Jingsen Zhang}, \bibinfo{person}{Zhiyuan Chen}, \bibinfo{person}{Jiakai Tang}, \bibinfo{person}{Xu Chen}, \bibinfo{person}{Yankai Lin}, \bibinfo{person}{Wayne~Xin Zhao}, \bibinfo{person}{Zhewei Wei}, {and} \bibinfo{person}{Ji-Rong Wen}.} \bibinfo{year}{2023}\natexlab{}.
\newblock \bibinfo{title}{A Survey on Large Language Model based Autonomous Agents}.
\newblock
\newblock
\showeprint[arxiv]{2308.11432}~[cs.AI]


\bibitem[Wang et~al\mbox{.}(2022)]%
        {wang2022self}
\bibfield{author}{\bibinfo{person}{Xuezhi Wang}, \bibinfo{person}{Jason Wei}, \bibinfo{person}{Dale Schuurmans}, \bibinfo{person}{Quoc Le}, \bibinfo{person}{Ed Chi}, \bibinfo{person}{Sharan Narang}, \bibinfo{person}{Aakanksha Chowdhery}, {and} \bibinfo{person}{Denny Zhou}.} \bibinfo{year}{2022}\natexlab{}.
\newblock \showarticletitle{Self-consistency improves chain of thought reasoning in language models}.
\newblock \bibinfo{journal}{\emph{arXiv preprint arXiv:2203.11171}} (\bibinfo{year}{2022}).
\newblock


\bibitem[Wei et~al\mbox{.}(2022)]%
        {wei2022chain}
\bibfield{author}{\bibinfo{person}{Jason Wei}, \bibinfo{person}{Xuezhi Wang}, \bibinfo{person}{Dale Schuurmans}, \bibinfo{person}{Maarten Bosma}, \bibinfo{person}{Fei Xia}, \bibinfo{person}{Ed Chi}, \bibinfo{person}{Quoc~V Le}, \bibinfo{person}{Denny Zhou}, {et~al\mbox{.}}} \bibinfo{year}{2022}\natexlab{}.
\newblock \showarticletitle{Chain-of-thought prompting elicits reasoning in large language models}.
\newblock \bibinfo{journal}{\emph{Advances in neural information processing systems}}  \bibinfo{volume}{35} (\bibinfo{year}{2022}), \bibinfo{pages}{24824--24837}.
\newblock


\bibitem[Williams(1992)]%
        {williams1992simple}
\bibfield{author}{\bibinfo{person}{Ronald~J Williams}.} \bibinfo{year}{1992}\natexlab{}.
\newblock \showarticletitle{Simple statistical gradient-following algorithms for connectionist reinforcement learning}.
\newblock \bibinfo{journal}{\emph{Machine learning}}  \bibinfo{volume}{8} (\bibinfo{year}{1992}), \bibinfo{pages}{229--256}.
\newblock


\bibitem[Wu et~al\mbox{.}(2020)]%
        {wu2020visual}
\bibfield{author}{\bibinfo{person}{Bichen Wu}, \bibinfo{person}{Chenfeng Xu}, \bibinfo{person}{Xiaoliang Dai}, \bibinfo{person}{Alvin Wan}, \bibinfo{person}{Peizhao Zhang}, \bibinfo{person}{Zhicheng Yan}, \bibinfo{person}{Masayoshi Tomizuka}, \bibinfo{person}{Joseph Gonzalez}, \bibinfo{person}{Kurt Keutzer}, {and} \bibinfo{person}{Peter Vajda}.} \bibinfo{year}{2020}\natexlab{}.
\newblock \bibinfo{title}{Visual Transformers: Token-based Image Representation and Processing for Computer Vision}.
\newblock
\newblock
\showeprint[arxiv]{2006.03677}~[cs.CV]


\bibitem[Wu et~al\mbox{.}(2023)]%
        {wu2023autogen}
\bibfield{author}{\bibinfo{person}{Qingyun Wu}, \bibinfo{person}{Gagan Bansal}, \bibinfo{person}{Jieyu Zhang}, \bibinfo{person}{Yiran Wu}, \bibinfo{person}{Beibin Li}, \bibinfo{person}{Erkang Zhu}, \bibinfo{person}{Li Jiang}, \bibinfo{person}{Xiaoyun Zhang}, \bibinfo{person}{Shaokun Zhang}, \bibinfo{person}{Jiale Liu}, \bibinfo{person}{Ahmed~Hassan Awadallah}, \bibinfo{person}{Ryen~W White}, \bibinfo{person}{Doug Burger}, {and} \bibinfo{person}{Chi Wang}.} \bibinfo{year}{2023}\natexlab{}.
\newblock \bibinfo{title}{AutoGen: Enabling Next-Gen LLM Applications via Multi-Agent Conversation}.
\newblock
\newblock
\showeprint[arxiv]{2308.08155}~[cs.AI]


\bibitem[Wu et~al\mbox{.}(2024)]%
        {wu2024stateflow}
\bibfield{author}{\bibinfo{person}{Yiran Wu}, \bibinfo{person}{Tianwei Yue}, \bibinfo{person}{Shaokun Zhang}, \bibinfo{person}{Chi Wang}, {and} \bibinfo{person}{Qingyun Wu}.} \bibinfo{year}{2024}\natexlab{}.
\newblock \bibinfo{title}{StateFlow: Enhancing LLM Task-Solving through State-Driven Workflows}.
\newblock
\newblock
\showeprint[arxiv]{2403.11322}~[cs.CL]


\bibitem[Xi et~al\mbox{.}(2023)]%
        {xi2023rise}
\bibfield{author}{\bibinfo{person}{Zhiheng Xi}, \bibinfo{person}{Wenxiang Chen}, \bibinfo{person}{Xin Guo}, \bibinfo{person}{Wei He}, \bibinfo{person}{Yiwen Ding}, \bibinfo{person}{Boyang Hong}, \bibinfo{person}{Ming Zhang}, \bibinfo{person}{Junzhe Wang}, \bibinfo{person}{Senjie Jin}, \bibinfo{person}{Enyu Zhou}, \bibinfo{person}{Rui Zheng}, \bibinfo{person}{Xiaoran Fan}, \bibinfo{person}{Xiao Wang}, \bibinfo{person}{Limao Xiong}, \bibinfo{person}{Yuhao Zhou}, \bibinfo{person}{Weiran Wang}, \bibinfo{person}{Changhao Jiang}, \bibinfo{person}{Yicheng Zou}, \bibinfo{person}{Xiangyang Liu}, \bibinfo{person}{Zhangyue Yin}, \bibinfo{person}{Shihan Dou}, \bibinfo{person}{Rongxiang Weng}, \bibinfo{person}{Wensen Cheng}, \bibinfo{person}{Qi Zhang}, \bibinfo{person}{Wenjuan Qin}, \bibinfo{person}{Yongyan Zheng}, \bibinfo{person}{Xipeng Qiu}, \bibinfo{person}{Xuanjing Huang}, {and} \bibinfo{person}{Tao Gui}.} \bibinfo{year}{2023}\natexlab{}.
\newblock \bibinfo{title}{The Rise and Potential of Large Language Model Based Agents: A Survey}.
\newblock
\newblock
\showeprint[arxiv]{2309.07864}~[cs.AI]


\bibitem[Xie et~al\mbox{.}(2024)]%
        {xie2024travelplanner}
\bibfield{author}{\bibinfo{person}{Jian Xie}, \bibinfo{person}{Kai Zhang}, \bibinfo{person}{Jiangjie Chen}, \bibinfo{person}{Tinghui Zhu}, \bibinfo{person}{Renze Lou}, \bibinfo{person}{Yuandong Tian}, \bibinfo{person}{Yanghua Xiao}, {and} \bibinfo{person}{Yu Su}.} \bibinfo{year}{2024}\natexlab{}.
\newblock \showarticletitle{Travelplanner: A benchmark for real-world planning with language agents}.
\newblock \bibinfo{journal}{\emph{arXiv preprint arXiv:2402.01622}} (\bibinfo{year}{2024}).
\newblock


\bibitem[Xu et~al\mbox{.}(2024)]%
        {xu2024core}
\bibfield{author}{\bibinfo{person}{Shuyuan Xu}, \bibinfo{person}{Zelong Li}, \bibinfo{person}{Kai Mei}, {and} \bibinfo{person}{Yongfeng Zhang}.} \bibinfo{year}{2024}\natexlab{}.
\newblock \bibinfo{title}{CoRE: LLM as Interpreter for Natural Language Programming, Pseudo-Code Programming, and Flow Programming of AI Agents}.
\newblock
\newblock
\showeprint[arxiv]{2405.06907}~[cs.CL]


\bibitem[Yao et~al\mbox{.}(2018)]%
        {yao2018taking}
\bibfield{author}{\bibinfo{person}{Quanming Yao}, \bibinfo{person}{Mengshuo Wang}, \bibinfo{person}{Yuqiang Chen}, \bibinfo{person}{Wenyuan Dai}, \bibinfo{person}{Yu-Feng Li}, \bibinfo{person}{Wei-Wei Tu}, \bibinfo{person}{Qiang Yang}, {and} \bibinfo{person}{Yang Yu}.} \bibinfo{year}{2018}\natexlab{}.
\newblock \showarticletitle{Taking human out of learning applications: A survey on automated machine learning}.
\newblock \bibinfo{journal}{\emph{arXiv preprint arXiv:1810.13306}} (\bibinfo{year}{2018}).
\newblock


\bibitem[Yao et~al\mbox{.}(2024)]%
        {yao2024tree}
\bibfield{author}{\bibinfo{person}{Shunyu Yao}, \bibinfo{person}{Dian Yu}, \bibinfo{person}{Jeffrey Zhao}, \bibinfo{person}{Izhak Shafran}, \bibinfo{person}{Tom Griffiths}, \bibinfo{person}{Yuan Cao}, {and} \bibinfo{person}{Karthik Narasimhan}.} \bibinfo{year}{2024}\natexlab{}.
\newblock \showarticletitle{Tree of thoughts: Deliberate problem solving with large language models}.
\newblock \bibinfo{journal}{\emph{Advances in Neural Information Processing Systems}}  \bibinfo{volume}{36} (\bibinfo{year}{2024}).
\newblock


\bibitem[Yao et~al\mbox{.}(2022)]%
        {yao2022react}
\bibfield{author}{\bibinfo{person}{Shunyu Yao}, \bibinfo{person}{Jeffrey Zhao}, \bibinfo{person}{Dian Yu}, \bibinfo{person}{Nan Du}, \bibinfo{person}{Izhak Shafran}, \bibinfo{person}{Karthik Narasimhan}, {and} \bibinfo{person}{Yuan Cao}.} \bibinfo{year}{2022}\natexlab{}.
\newblock \showarticletitle{React: Synergizing reasoning and acting in language models}.
\newblock \bibinfo{journal}{\emph{arXiv preprint arXiv:2210.03629}} (\bibinfo{year}{2022}).
\newblock


\bibitem[Yao et~al\mbox{.}(2023)]%
        {yao2023react}
\bibfield{author}{\bibinfo{person}{Shunyu Yao}, \bibinfo{person}{Jeffrey Zhao}, \bibinfo{person}{Dian Yu}, \bibinfo{person}{Nan Du}, \bibinfo{person}{Izhak Shafran}, \bibinfo{person}{Karthik Narasimhan}, {and} \bibinfo{person}{Yuan Cao}.} \bibinfo{year}{2023}\natexlab{}.
\newblock \showarticletitle{{ReAct}: Synergizing Reasoning and Acting in Language Models}. In \bibinfo{booktitle}{\emph{International Conference on Learning Representations (ICLR)}}.
\newblock


\bibitem[Yuan et~al\mbox{.}(2024)]%
        {yuan2024easytool}
\bibfield{author}{\bibinfo{person}{Siyu Yuan}, \bibinfo{person}{Kaitao Song}, \bibinfo{person}{Jiangjie Chen}, \bibinfo{person}{Xu Tan}, \bibinfo{person}{Yongliang Shen}, \bibinfo{person}{Ren Kan}, \bibinfo{person}{Dongsheng Li}, {and} \bibinfo{person}{Deqing Yang}.} \bibinfo{year}{2024}\natexlab{}.
\newblock \showarticletitle{EASYTOOL: Enhancing LLM-based Agents with Concise Tool Instruction}.
\newblock \bibinfo{journal}{\emph{arXiv preprint arXiv:2401.06201}} (\bibinfo{year}{2024}).
\newblock


\bibitem[Zeng et~al\mbox{.}(2024)]%
        {zeng2024flowmind}
\bibfield{author}{\bibinfo{person}{Zhen Zeng}, \bibinfo{person}{William Watson}, \bibinfo{person}{Nicole Cho}, \bibinfo{person}{Saba Rahimi}, \bibinfo{person}{Shayleen Reynolds}, \bibinfo{person}{Tucker Balch}, {and} \bibinfo{person}{Manuela Veloso}.} \bibinfo{year}{2024}\natexlab{}.
\newblock \bibinfo{title}{FlowMind: Automatic Workflow Generation with LLMs}.
\newblock
\newblock
\showeprint[arxiv]{2404.13050}~[cs.CL]


\bibitem[Zhang et~al\mbox{.}(2020)]%
        {bert-score}
\bibfield{author}{\bibinfo{person}{Tianyi Zhang}, \bibinfo{person}{Varsha Kishore}, \bibinfo{person}{Felix Wu}, \bibinfo{person}{Kilian~Q. Weinberger}, {and} \bibinfo{person}{Yoav Artzi}.} \bibinfo{year}{2020}\natexlab{}.
\newblock \bibinfo{title}{BERTScore: Evaluating Text Generation with BERT}.
\newblock
\newblock


\end{thebibliography}

\end{document}